\providecommand{\pgfsyspdfmark}[3]{}
\documentclass{esannV2}
\usepackage{graphicx}
\usepackage[latin1]{inputenc}
\usepackage{amssymb,amsmath,array}

\usepackage{subcaption}

\usepackage[a4paper]{geometry}
\usepackage{fancyhdr}
\begin{document}
\title{Bioinformatics and Medicine in the Era of Deep Learning}

\author{Davide Bacciu$^1$\thanks{D.Bacciu is funded by the Italian MIUR under project SIR-RBSI14STDE}, Paulo J.G. Lisboa$^2$, Jos\'{e} D. Mart\'{i}n$^3$,\\ Ruxandra Stoean$^4$, Alfredo Vellido$^5$\thanks{A. Vellido is funded by the Spanish MINECO under project TIN2016-79576-R}
%
\vspace{.3cm}\\
%
\vspace{.1cm}
1- Universit\`a di Pisa - Italy\\
\vspace{.1cm}
2- Liverpool John Moores University - UK\\
\vspace{.1cm}
3- Universitat de Val\`{e}ncia - Spain\\
\vspace{.1cm}
4- University of Craiova - Romania\\
\vspace{.1cm}
5- Universidad Polit\'{e}cnica de Catalu\~{n}a - Spain
}

\maketitle
\thispagestyle{fancy}

\begin{abstract}
Many of the current scientific advances in the life sciences have their origin in the intensive use of data for knowledge discovery. In no area this is so clear as in bioinformatics, led by technological breakthroughs in data acquisition technologies. It has been argued that bioinformatics could quickly become the field of research generating the largest data repositories, beating other data-intensive areas such as high-energy physics or astroinformatics. Over the last decade, deep learning has become a disruptive advance in machine learning, giving new live to the long-standing connectionist paradigm in artificial intelligence. Deep learning methods are ideally suited to large-scale data and, therefore, they should be ideally suited to knowledge discovery in bioinformatics and biomedicine at large. In this brief paper, we review key aspects of the application of deep learning in bioinformatics and medicine, drawing from the themes covered by the contributions to an ESANN 2018 special session devoted to this topic.
\end{abstract}

\section{Introduction}
\begin{figure}
\centering
\includegraphics[width=.7\textwidth]{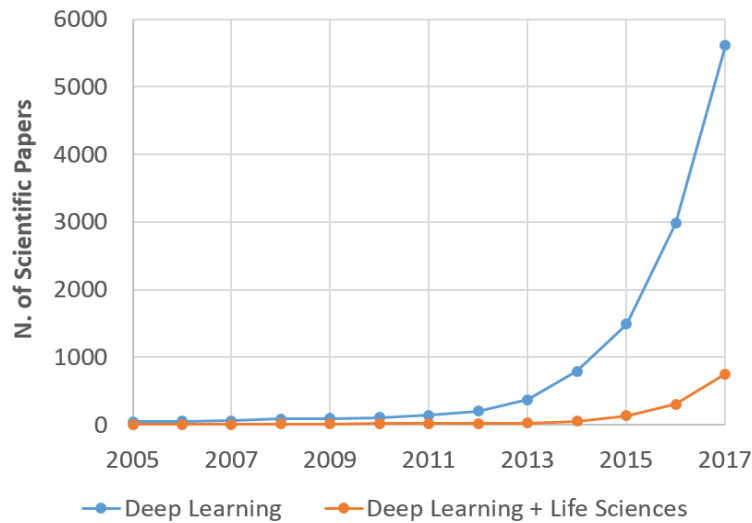}
\caption{Evolution on the number of papers published on DL topics with respect to those on DL in BioInformatics (Source: query on abstracts keywords on Scopus data; date: February 1$^{st}$, 2018)}. \label{fig:papers}
\end{figure}

Deep Learning (DL) \cite{Goodfellow-et-al-2016} has become an increasingly popular Machine Learning (ML) approach in the last decade, as shown in Fig.~\ref{fig:papers}. One of the main reasons for its success stems from its internal representation in the form of high-level features, allowing the modelling of difficult problems, and a smart initialization of some other deep structures. Moreover, staging the difficult task of efficient feature selection by using multiple layers has been crucial in solving extremely difficult problems of image classification 
\cite{NIPS2012_4824,brusilovsky:simonyan2014very,NIPS2015_5638}, or Natural Language Processing (NLP) 
\cite{graves2013speech,Sutskever:2014,DBLP:journals/corr/NallapatiXZ16}
by means of Convolutional Neural Networks (CNNs) and Deep Recurrent Neural Networks (RNN), respectively. The success achieved in such complicated problems has generated a great interest not only in the academic community but also in industry, with many private companies involved in the development of commercial products based on DL \cite{Seonwoo_2017, forbes_murnane}.

Although CNNs and recurrent networks have already produced significant advances in biomedical imaging and biomedical signal processing \cite{Seonwoo_2017}, the impact of DL in Bioinformatics is still limited. This is possibly related to
some of the challenges that many bioinformatics data sets represent, such as insufficient and unbalanced data, or challenges related to the area of application itself, such as the lack a straightforward interpretation of deep models. This special session aims to bring together some of the most recent advances in DL as applied to bioinformatics and (bio)medicine.

The contents of this tutorial are outlined as follows: Section \ref{sec:structured_biomedical} deals with the problem of structured biomedical data. Section \ref{sec:image_processing} reviews the contribution of DL to image processing in medicine. Section \ref{interp} provides a glimpse of a key issue when dealing with DL approaches in the clinical field, namely, intepretability. Finally, Section \ref{sec:future} gives some insights on future trends and challenges derived from the papers submitted to the session.

\section{Deep Learning of structured biomedical data}
\label{sec:structured_biomedical}

The term structured data denotes a whole selection of data types encoding different forms of relational information, ranging from sequences to more complex and general classes of graphs. Structured data can be seen as representing compound information made of atomic entities, i.e. the labels of the nodes/vertices of the structures, linked by relationships, denoted by the edges between the nodes (possibly complemented by edge labels). As such, structured data place emphasis on evaluating and interpreting information ``in context'' rather than following the classical i.i.d. assumption, where the context is defined by the set of nodes an atomic piece of information is linked to.

Structured data arise naturally as the result of many biological, physical and chemical processes. Sequences are the simplest form of structured data and are a natural representation, for instance, for nucleotide compounds (e.g. DNA, RNA) and for physiological signals (e.g. ECG, EEG, MEG data). Trees, instead, are widely used to represent cell phylogeny and clonal diversity, while more general classes of graphs find wide application to represent proteomic and metabolomic interactions, protein structures as well as chemical compounds.

The field of ML for structured data processing has a long consolidated history dating back to the sixties for the first approach to sequential data, namely the Hidden Markov Model (HMM) \cite{baum1966}. Later, in the nineties, the scientific community started tackling more complex structures, with works proposing recursive approaches for acyclic graph processing \cite{treeFram98,bhtmm}, later followed by a number of approaches tackling more general classes of graphs with cycles \cite{scarselli09,nn4g}.

The first and most popular applications of DL to biomedicine concern the processing of sequential data. CNN models have found wide application to genomics, in particular for the prediction of protein binding sites such as in the DeepBind \cite{DeepBind} and DeepSEA \cite{deepsea} methods, where it has also been proposed as a method to visualize the effect of wild-type mutation on binding site prediction, contributing to the interpretability of the learned model.  Despite their limitation in treating only fixed length subsequences, CNNs have quickly become the reference DL model for genomic studies: a recent review of the main CNN approaches in the field can be found in \cite{Jones257}.  RNNs, such as the Long Short Term Memory (LSTM), have found less application in genomics despite their ability in modeling variable length sequences, mostly due to a widespread belief of the biomedical community, which considers them difficult to train \cite{Angermueller878}.  An attempt to bring the LSTM model into the biomedical community is put forward in \cite{distill}, where an approach to learn interpretable features from an LSTM trained on real-world clinical time-series is proposed. Along the same line, \cite{lipton} proposed the use of a LSTM to classify $128$ diagnoses from multivariate clinical time series collected in an intensive care unit (ICU). Use of more parsimonious RNN models is discussed in \cite{neurocomp2017} for scoring stress levels from heart-rate information. On the unsupervised learning side, stacked denoising autoencoders have been proposed to learn informative encodings on noisy ECG time series \cite{sdaECG}. The ubiquitous Generative Adversarial Networks (GANs) have very recently been used \cite{organ} for the generation of molecules encoded as character sequences.

Several papers of this ESANN 2018 special session deal with DL approaches for biomedical sequences. In \cite{bianchi}, it is proposed a framework integrating deep autoencoders with kernel methods to learn effective encodings of multivariate clinical time series in presence of missing information. A deep RNN approach is discussed in \cite{deepESN} to diagnose Parkinson's disease from spiral drawing tests. CNNs are used in \cite{cnnEEG} to classify sleep stages from EEG recordings, using a fixed window approach splitting time series into subsequences of 30 seconds of electrical activity.

The long wave of the DL revolution is now approaching and rediscovering the processing of more complex and expressive forms of structured data, in particular as regards cheminformatics and drug discovery. Again, CNNs are among the most popular approaches here: \cite{cnnFing} proposed the first attempt to process molecular data within a CNN using circular fingerprint encodings imported from the chemistry literature. A proper generalization of the concept of convolution from images to graphs of varying dimensionality has been proposed in \cite{cnnpatch}, with application to the characterization of chemical compounds. Graph convolutions are used in \cite{cnnLSTM} together with an attentional model based on LSTM to predict molecule properties with a one-shot learning approach. Very recently, a DL model, based on a similar encoding to the one in \cite{nn4g}, has been proposed in \cite{Jure17} for the processing of large scale networks of proteomics and metabolomics interactions for predicting new drug-disease associations in drug repurposing. The special session contributes to this flourishing topic with a paper \cite{deepSP} discussing a deep graph kernel for disease gene prioritization in bioinformatics.

\section{Deep Learning for medical image processing}
\label{sec:image_processing}

The popularity of DL has exponentially risen especially due to its capability to process images independently from human intervention. This includes inherent robustness to variations in position, rotation, scale, perspective and occlusion. These traits have consequently appeared as particularly valuable in the medical sector. The amount of image data available for analysis keeps increasing with the modernization and constant use of imaging devices. Time-saving decision support in this area was achieved by ML techniques before the arrival of DL, but with a different supplementary human cost, i.e. that of highlighting the regions of interest in the images, of handcrafting the relevant features for the diagnosis and of labelling each image. This has proved to be a turning point for DL, as the accurate provider of both image processing and image interpretation. 
Lacking the need for expert handcrafted features by automatically learning the optimal attributes from the available images \cite{litjens} and benefiting from large amounts of available data, DL has therefore successfully entered the realm of medical imaging. Its applications range from landmark detection and tissue segmentation to diagnosis and prognosis \cite{shen}.

The main DL approach to image analysis is the CNN. Its special convolutional layers are able to detect the relevant features gradually, from those low-level to the high-level structures, by inspecting small portions of the training images. Following the independent feature discovery, the image is labelled according to the given task. CNN modelling may target images from histopathology 
\cite{albarqouni,stoean}, CT scans \cite{anthimopoulos,roth}, or MRI \cite{pereira}, to name a few. Stacked autoencoders can also be employed for automatic feature extraction, such as in \cite{wu} for MRI imaging and in \cite{cheng} for CT and ultrasound images.

The success of DL for medical imaging can be appreciated in the constant presence of the topic at conferences specialized in biomedical computing 
and in journals dealing with medical image processing, 
as well as in the advent of companies using these methods in the area (e.g. Enlitic, Arterys, or Lunit \cite{DLAMI}).


Shifting now from the brighter side of the DL potential for image mining in medicine, a big problem that the paradigm faces in this real-world complex field of application, as compared to its use for general images, must be acknowledged. The actual scarcity of labelled medical images makes it harder for the approach to perform well, leading to overfitting and hard parametrization. Current solutions include data geometric augmentation, transfer learning and fine-tuning \cite{shen,greenspan} from general image data sets such as ImageNet or preferably \cite{ravi17} from those emerging in the medical domain (e.g. on Kaggle), as well as GANs \cite{wolterink}.

Several papers in the current special session cover some of the key aspects of the application of DL for image processing in medicine and public health. As the field of application is concerned, two of them target medical imaging tasks from CT and HP and one deals with recognising pollen from microscopic images for allergy and asthma prevention in medicine. As for the DL architectures involved, two CNN (of which one is a GAN) and one autoencoder emulator are employed to solve the problems.
In \cite{GANESANN}, the authors propose an expansion of an initially low number of available 2D lung CT scans through a GAN methodology. In \cite{NMFESANN}, a NMF approach for learning a reduced feature representation in a fashion similar to an autoencoder is described for the processing of histopathological colorectal cancer slides. Finally, from the public health perspective, a CNN for pollen recognition from a collection of microscopic images is employed in \cite{pollen}.

\section{Interpretable DL models in biomedicine and bioinformatics}
\label{interp}

Over the last decades, data have become central to the life sciences. Biomedicine and, perhaps more expressively, bioinformatics are perfect examples of that. The introduction of computerized and networked digital systems and the impressive advances in non-invasive data acquition technologies are putting data at the heart of these disciplines.

Data are hoped to become the key to the discovery of new knowledge at all physiological scales, opening the doors to hitherto unaccessible medical advances. Such transformation from data to knowledge is a natural goal for ML. To be accepted in biomedicine and bioinformatics, and very especially in medical practice, ML-based approaches must be trusted. One of the main challenges ML faces to achieve such desired trustworthiness is that of becoming explainable and interpretable \cite{vellido_etal12,lisboa13}.

The relevance of this challenge is heightened by a pressing societal issue: the implementation of the new European Union directive for General Data Protection Regulation (GDPR). It is to be enforced throughout 2018 with minimum variation between European countries. Its Article 22, concerning ``Automated individual decision-making, including profiling'', mandates a \textit{right to explanation} of all decisions made by automated or artificially intelligent algorithmic systems \cite{goodman17}. The GDPR directive makes model interpretability a core concern in biomedical decison making; it directly involves ML and should particularly concern those aiming to see DL being used in medical practice, beyond basic research.

Another of the reasons for explainability and interpretability to have become, of late, contentious and hotly debated issues in ML is, precisely, the new life given to connectionism in the form of DL. In general, DL approaches are extreme cases of \textit{black box} models. Being such a success story for ML, DL models' lack of interpretability has become a pressing concern in the area, reflected by recent literature. A concern of no easy solution, given the difficulty of making these often extremely complex systems become somehow transparent. In \cite{dong17}, an adversarial training scheme was recently proposed, where model neurons ``are endowed with human-interpretable concepts'' and interpretable representations can trace outcomes back to influential neurons, providing an explanation of how models make their predictions. Also recently, interpretability criteria based on analysis of deep networks in the information plane \cite{schwartz17} were described.

Lack of transparency has indeed been argued to be one of the main barriers to the acceptance and adoption in medicine of ML methods in general and DL methods in particular \cite{ravi17}. This view is shared by Che and co-workers; in \cite{che15}, they propose gradient boosting decision trees to extract interpretable knowledge from a trained deep network. In related work \cite{wu17}, deep models are regularized so that their class-probability predictions can be modeled with minimum loss by decision trees with few nodes. Interpretation from these trees is far more tractable and intuitive than from the original models.

An example of how to deal with interpretability in DL can be found in the ESANN 2018 special session covered in this tutorial. In \cite{amorimESANN18}, an extension of \textit{interpretable mimic learning} that teaches reasonably simple and interpretable models so as to mimic predictions of complex DL models without missing in the performance is introduced. It focuses on problems of ordinal classification and illustrates the capabilities of the model with a problem of ordinal response to cancer treatment.

\section{Some future trends and challenges}
\label{sec:future}

The application of DL methods to problems in biomedicine and bioinformatics is a many-faceted problem. At this point in time and in a brief tutorial such as this, it would be impossible to provide a comprehesive view of the future trends in the area and the various important challenges faced by such applications in real-world scenarios.

Some future trends, though, have been outlined for biomedicine in \cite{mamoshina16} and for bioinformatics in \cite{Seonwoo_2017}. For biomedicine, DNNs are quoted to be of potential benefit to areas as varied as semantic linking, biomarker development, drug discovery (structural analysis and hypothesis formulation through DNN abstract learned representations analysis), clinical recommendations and transcriptomic data analysis
. For bioinformatics, investigating proper ways to encode raw
and multi-modal data forms, instead of human-processed features, and learn suitable features from those multi-modal or raw forms is seen as a future challenge for DNNs.

Note that the previous comments do not cover medical applications in clinical settings and healthcare. In these, and as reported in the previous sections, explainability and interpretability are major challenges for DL.
Despite extensive research into the interpretation of neural networks in clinical settings dating back several decades \cite{YANG1998917}  and extending to more recent work \cite{STURM2016141}, we are only now weaving the first strands of methods capable of \textit{translating} the complex inner workings of deep architectures. This means that this is possibly both a future strong research trend and far-from-trivial challenge. The latter is exemplified by a recent study by Google Brain researchers \cite{adebayo18}, who show the surprising result of the lack of sensitivity of local model explanations to deep neural network (DNN) parameter values. The study concludes that ``the architecture of DNN is a strong prior on the input, and with random initialization, is able to capture low-level input [image] structure''.

As mentioned, another likely future trend and definite challenge for DL is the coherent integration of multi-modal data \cite{bhanot17}. A case in multi-omics data integration is presented in one of the studies in the ESANN 2018 special session covered by this tutorial \cite{bica18}. It delivers a novel \textit{super-layered} neural network architecture named cross-modal neural network. Interestingly, it is meant to perform well in scenarios with a limited number of training examples available (something not uncommon in multi-modal biomedical problems).

A final glimpse of the future is provided in \cite{wulfing18}, where a reinforcement learning brain-dynamics-model-free approach to control response properties of \textit{biological} neural networks is described. Beyond this proof-of-concept, such approach could have an enormous impact on the study of degenerative brain diseases.


\begin{footnotesize}



\end{footnotesize}


\end{document}